\documentclass{bmvc2k}
\usepackage{hyperref}
\usepackage{multirow}
\usepackage{amsmath}
\usepackage{graphicx}
\usepackage{marginnote}
\title{Pedestrian Action Anticipation using Contextual Feature Fusion in Stacked RNNs}

\addauthor{Amir Rasouli}{http://www.eecs.yorku.ca/~aras/}{1}
\addauthor{Iuliia Kotseruba}{http://www.eecs.yorku.ca/~yulia_k/}{1}
\addauthor{John K. Tsotsos}{http://www.eecs.yorku.ca/~tsotsos/Tsotsos/Home.html}{1}

\addinstitution{
 Dept. of Electrical Engineering and Computer Science\\
 York University\\
 Toronto, Canada
}

\runninghead{A.Rasouli, I. Kotseruba, J. Tsotsos}{Pedestrian Action Anticipation}

\begin{document}

\maketitle
\marginnote{\rotatebox{90}{This work was accepted and presented at BMVC 2019}}[0.50cm]
\begin{abstract}
One of the major challenges for autonomous vehicles in urban environments is to understand and predict other road users' actions, in particular, pedestrians at the point of crossing. The common approach to solving this problem is to use the motion history of the agents to predict their future trajectories. However, pedestrians exhibit highly variable actions most of which cannot be understood without visual observation of the pedestrians themselves and their surroundings. To this end, we propose a solution for the problem of pedestrian action anticipation at the point of crossing. Our approach uses a novel stacked RNN architecture in which information collected from various sources, both scene dynamics and visual features, is gradually fused into the network at different levels of processing. We show, via extensive empirical evaluations, that the proposed algorithm achieves a higher prediction accuracy compared to alternative recurrent network architectures. We conduct experiments to investigate the impact of the length of observation, time to event and types of features on the performance of the proposed method. Finally, we demonstrate how different data fusion strategies impact prediction accuracy. 
\end{abstract}

\section{Introduction}
\label{sec:intro}

Autonomous driving systems suitable for urban environments require the ability to comprehend and anticipate the actions of other road users. In this context, pedestrians are of particular importance being the most vulnerable road users, especially when crossing the road. Anticipating pedestrian crossing action helps the driving systems to select the correct course of action to avoid any potential collisions and disruption of traffic flow. 
  
\begin{figure}[!tb]
\centering
\includegraphics[width=1\textwidth]{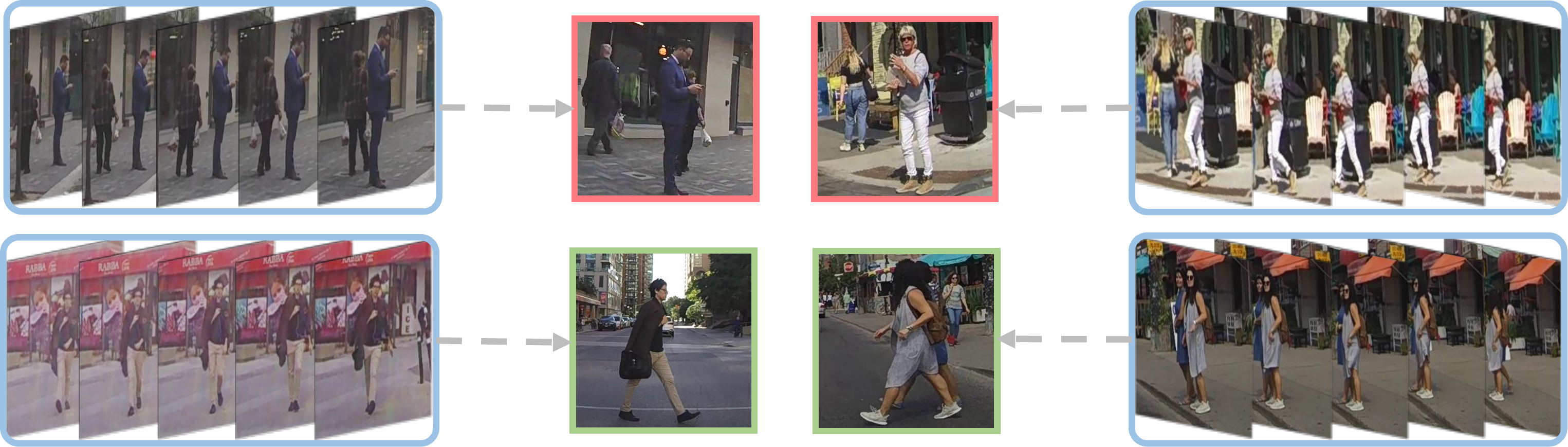}
\caption{Examples of pedestrians prior to making crossing decision. Green and red colors indicate whether the pedestrian will or will not cross.}
\label{pedestrians_crossing}
\end{figure}

Today, the dominant approaches to solving the problem of pedestrian action prediction are trajectory-based. These algorithms rely on the motion patterns of pedestrians and predict their trajectories at  some time in the future \cite{kooij2014context,Alahi2016social,bhattacharyya2018long}. Although pedestrian dynamics are important, they are not sufficient for making sense of pedestrian behavior and predicting  their actions as they are often subject to error. For example, a pedestrian intending to cross the street could be standing at the intersection (with no motion history), walking alongside the road or abruptly changing their walking pattern prior to crossing \cite{Schmidt2009} (see Figure \ref{pedestrians_crossing}). In addition, pedestrians exhibit highly variable motion patterns which can be influenced by various environmental factors such as signals, the ego-vehicle motion, road structure, etc.  All of these factors add to the complexity of predicting pedestrian actions \cite{rasouli2017agreeing}. Thus a statistical inference on pedestrian trajectories alone may not sufficient for predicting their actions. To remedy this problem, some algorithms, in addition to pedestrians' trajectories, use information such as head orientation \cite{kooij2014context}, social interactions \citep{Alahi2016social,gupta2018social} or destination locations \cite{Bai2015,rehder2018pedestrian} to predict  pedestrians' forthcoming actions. However, to achieve a more robust prediction in a complex traffic scene, there is a need for a more general approach that exploits various sources of contextual information. 

To this end, we propose a novel algorithm for pedestrian action anticipation in video sequences recorded with an on-board camera inside a moving vehicle. Given that the main point of interaction between autonomous vehicles and pedestrians is at the time of crossing, here we particularly focus on the pedestrian crossing anticipation, i.e. we determine whether an observed pedestrian will cross in front of the vehicle. For this purpose, instead of focusing on a subset of contextual elements,  we take a more holistic approach and combine information from multiple sources, including the appearance, pose and dynamics of pedestrians,their surrounding context, and the ego-vehicle speed. We introduce a novel stacked recurrent neural network architecture in which data from different modalities  are gradually fused at different levels of processing. We conduct experiments to evaluate the performance of the proposed algorithm and discuss how different data arrangement strategies affects its performance.

\section{Related Works}

Action recognition and prediction are widely studied topics in the field of computer vision \cite{Yoo_2016_CVPR, Alahi2016social,carreira2017quo,Mahmud_2017_ICCV,Lee_2017_CVPR,Chen_2018_ECCV,Oliu_2018_ECCV,
Suzuki_2018_CVPR}. Among commonly used methods for modeling sequential data are recurrent neural networks (RNNs). In particular, their variants, Gated Recurrent Units (GRUs) \cite{chung2014empirical} and Long Short-Term Memory (LSTM) units \cite{hochreiter1997long} have a wide range of applications for activity recognition \cite{veeriah2015differential,Singh2016}, action prediction \cite{Alahi2016social, Mahmud_2017_ICCV} and video captioning \cite{baraldi2017hierarchical, jiang2018recurrent}. In their simplest form, RNNs are used in a single-layer format \cite{veeriah2015differential, Lee_2017_CVPR, Mahmud_2017_ICCV, jiang2018recurrent} for temporal reasoning with the objective of classifying activities or predicting future behavior. RNNs are also used in multi-stream architectures \cite{Singh2016,gammulle2017two, bhattacharyya2018long} in which, for example, data of different modalities is processed in separate streams and combined at the end for inference \cite{bhattacharyya2018long}. Some approaches increase the depth of RNNs in space by stacking them on top of each other \cite{pascanu2013construct,donahue2015long, yue2015video,ilyes2018residual,liu2018skeleton}, so that the input to the RNN at each level consists of the hidden states of the RNN in the previous layer. A variation of this is stacked RNNs that are organized in a hierarchy  \cite{du2015hierarchical, yu2016video, pan2016hierarchical, baraldi2017hierarchical,li2017adaptive} where the length of  information flow  in each successive layer is reduced to minimize the computation cost. In both stacked and hierarchical architectures, the raw inputs, e.g. image features, enter the network at the bottom layer, and the representations of the first layer are propagated through the network.   

RNN-based architectures show particularly  promising results in action prediction applications such as future scene generation \cite{Liang_2017_ICCV, Oliu_2018_ECCV,Byeon_2018_ECCV}, activity class prediction \cite{donahue2015long,jain2016recurrent, Mahmud_2017_ICCV}, event anticipation \cite{chan2016anticipating,zeng2017agent,Suzuki_2018_CVPR}, and trajectory estimation \cite{Alahi2016social,
Lee_2017_CVPR, bhattacharyya2018long}. The latter is the most commonly used in the field of pedestrian action prediction in natural scenes. These techniques often use encoder-decoder architectures in which one RNN unit is responsible for encoding observations, often by tracking the locations of pedestrians,  and generating representations which are used by a decoder, often another RNN unit, to infer future predictions.

Motion histories are often augmented with additional contextual information to achieve more robust predictions. For example, in \cite{Alahi2016social,robicquet2016learning,gupta2018social} the social interactions between agents are taken into account for forecasting their future trajectories. These approaches, however, are applied to top-down view scenes (recorded with stationary camera) where complex social interactions can be observed easily.  In the context of driving, action prediction is more challenging due to the ego-vehicle motion which may also impact the behavior of other agents. In \cite{bhattacharyya2018long} the authors use a two-stream encoder-decoder LSTM model which simultaneously predicts pedestrian trajectory and future speed of the ego-vehicle. The authors combine the estimated vehicle speed with the representations of pedestrians' trajectories to predict their future motion. Although scene information is used for speed estimation, no visual information involving the pedestrians, aside from their trajectories, is used for inference. In some works, in addition to scene dynamics, authors use cues such as pedestrian's head orientation and proximity to the curb \cite{kooij2014context} or pedestrian's potential destination locations \cite{Bai2015,rehder2018pedestrian}, however, these approaches still predominantly rely on pedestrian's trajectories which might not be sufficient for the reasons mentioned earlier.

Some algorithms address pedestrian action prediction as a classification problem and use various contextual elements such as road structure \cite{2014_bonin, Schneemann2016},  head orientation and signals \cite{Rasouli_2017_ICCV_Workshops} to predict pedestrian intention of crossing. These models, however, do not take into account all information relevant for reasoning, e.g. scene dynamics \cite{Rasouli_2017_ICCV_Workshops}, do not propose a mechanism for visual processing of the scenes \cite{Schneemann2016} or are evaluated in a very limited context \cite{2014_bonin}.

\textbf{Contributions.} This paper contributes the followings: 1) We propose a novel stacked recurrent network architecture with multilevel feature fusion for predicting pedestrian crossing action. The proposed algorithm benefits from a combination of various visual and motion features for prediction\footnote{The code is available at \url{https://github.com/aras62/SF-GRU}}. 2) We perform extensive experiments to evaluate the performance of the proposed algorithm against alternative RNN architectures. 3) We examine the effects of time to event and observation duration on the accuracy of crossing prediction. 4) We investigate the influence of various sources of contextual information on the performance of the proposed algorithm. 5) In the end, we show how changing the order of feature inputs during training and inference according to their level of complexity affects prediction.

\section{Approach}
We define pedestrian crossing prediction as a binary classification problem in which the objective is to determine whether a pedestrian $i$ will cross the street given the observed context up to some time $m$. The prediction relies on five sources of information including the local context $\{C_{p_i},C_{s_i}\}$, where $C_{p_i} = \{c^1_{p_i},...,c^m_{p_i}\} $ and $C_{s_i} = \{c^1_{s_i},...,c^m_{s_i}\} $ refer to visual features of the pedestrian and their surroundings respectively, pedestrian pose $P_i = \{p^1_i,...,p^m_i\}$, 2D bounding box locations $B_i = \{ b^1_i,...,b^m_i \}$, where $b_i$ is a two-point coordinate $[(x1_i,y1_i)(x2_i,y2_i)]$ corresponding to the top-left and bottom-corner of the bounding box around the pedestrian, and the speed of the ego-vehicle $S = \{s^1,...,s^m\}$.

\begin{figure}[!tb]
\centering
\includegraphics[width=1\textwidth]{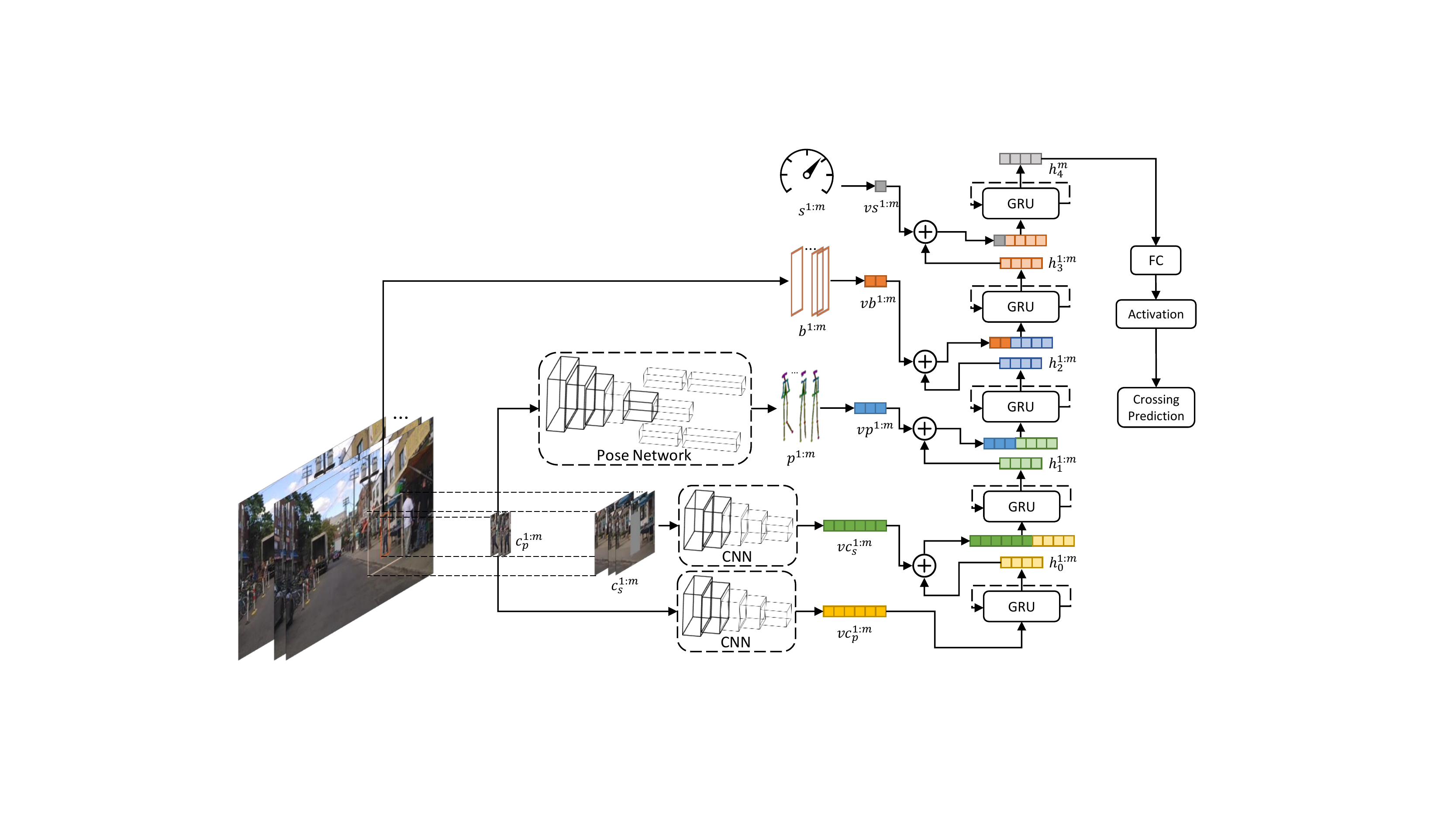}
\caption{The architecture of the proposed algorithm \textit{SF-GRU} comprised of five GRUs each of which processes a concatenation of features of different modalities and the hidden states of the GRU in the previous level. The information is fused into the network gradually according to the complexity of the features. Each feature input consists of $m$ sequential observations. From bottom to top layers features are fused as follows:  pedestrian appearance $c_p^{1:m}$, surrounding context $c_s^{1:m}$, poses $p^{1:m}$, bounding boxes $b^{1:m}$ and ego-vehicle speed $s^{1:m}$. $\bigcirc$\kern-0.88em{$+$} refers to concatenation operation.  }
\label{architecture}
\end{figure} 

\subsection{Architecture}
Recurrent neural networks (RNNs) are extensions of feedforward networks. RNNs have recurrent hidden states allowing them to learn temporal dependencies in sequence data. This inherent temporal depth has been shown to greatly benefit tasks, such as pedestrian trajectory prediction, that apply single-layer RNNs to point coordinates in a space. In addition to temporal depth, spatial depth of RNNs can also be increased by stacking multiple layers of RNN units on top of one another. This approach is an effective way of improving sequential data modeling in complex tasks \cite{levine2018benefits}, in particular, video sequence analysis \cite{donahue2015long,yue2015video} in which the network models dependencies between visual features of consecutive video frames. 

Given the multimodal nature of pedestrian action anticipation which relies on both dynamics and visual scene information, we employ a hybrid approach. We use a stacked RNN architecture similar to \cite{yue2015video} in which we gradually fuse the features at each level according to their complexity. In other words, we input the visual features of the scene that can benefit more from spatial depth of the network at the bottom layers and the dynamics features, e.g. trajectories and speed, at the higher levels of the network (see Figure \ref{architecture}). Below, we describe the procedures to generate each data type we use in the proposed model.  

\textbf{Local context.}  At each time step of the observation, for each pedestrian, we use their appearance and surroundings. The former is captured using images cropped to the size of 2D bounding boxes around the pedestrian in the frame. For the surroundings,  we extract a region around the pedestrian by scaling up the 2D bounding box coordinates, and squarifying the dimensions so the width of the scaled bounding box matches its height. This gives us a wider viewing angle of the scene around the pedestrian which may include street, other pedestrians, signals or traffic. In the surround crop, we suppress the pedestrian appearance by setting the pixel values in the original bounding box coordinate to neutral gray. Both appearance and surround crops are processed  using a convolutional neural network (CNN) which produces two feature vectors $vc^{1:m}_p$ and $vc^{1:m}_s$.

\textbf{Pose.} A pose network is used to generate 18 body joints coordinates, each corresponding to a point in 2D space, for each pedestrian. The joint coordinates are normalized and concatenated into a 36D feature vector $vp^{1:m}$. 

\textbf{2D bounding box.} We transform the bounding box coordinates into relative displacement from the initial position forming a feature vector $vb^{1:m}$. This can be seen as the velocity of the pedestrian at every time step. 

\textbf{Speed.} Speed is a vector of the ego-vehicle speed recordings for each time step $vs^{1:m}$ in $km/h$.

\textbf{Multimodal feature fusion.} For the joint modeling of our sequence data, we use gated recurrent units (GRUs) \cite{chung2014empirical} which are simpler compared to LSTMs and, in our case, achieve similar performance.  Recalling the equation of GRU, the $j^{th}$ level of the stack is given by, 
\begin{equation}
\begin{gathered} 
r^t_j = \sigma(W^{xr}_jx^t_j+ W^{hr}_jh^{t-1}_j),\\
z^t_j = \sigma(W^{xz}_jx^t_j+ W^{hz}_jh^{t-1}_j),\\
\tilde{h}^t_j = tanh(W^{xh}_jx^t_j+ W^{hh}_j(r^t_j \odot h^{t-1}_j),\\
h^t_j = (1-z^t_j)\odot h^{t-1}_j + z^t_j \odot \tilde{h}^t_j),
\end{gathered}
\end{equation}

\noindent where $\sigma(.)$ is the sigmoid function,  $r^t$ and $z^t$ are reset and update gates, and matrices $W^{..}$ are weights between two units. For $j = 0$ (the bottom level of the stack), $x^t_0 = vc^t_p$ and for $j > 0$, $x^t_j = h^t_{j-1} + vy^t[j-1]$ where $y^t = \{vc^t_s, vp^t, vb^t, vs^t \}$. The final prediction is achieved by a linear transformation of $h^t_n$ where $n$ is the number of levels (in our case 5) in the proposed stacked architecture. In the training phase we use the \textit{binary cross-entropy} loss function.

\subsection{Implementation}
In our architecture, we use GRUs \cite{chung2014empirical} with 256 hidden units.  For local context, we crop the pedestrian samples $C_p$ using the 2D bounding box annotations, resize them so the larger dimension is equal to 224 and pad them with zeros to preserve the aspect ratio. For surround context, $C_s$, we use $1.5x$ (set empirically) scaled version of the 2D bounding boxes. The parts of the cropped images that include pedestrians of interest are suppressed by neutral gray with RGB of $(128,128,128)$. We resize these images to $224 \times 224$. The local context images are processed using VGG16 \cite{simonyan2014very} (without fully connected (\textit{fc}) layers) pretrained on ImageNet \cite{ILSVRC15} followed by an average global pooling generating a feature vector of size 512 per crop. For pedestrian poses, we use \cite{cao2017realtime} which is pretrained on the COCO dataset \cite{coco}. The network generates 18-joint pose per pedestrian sample.

\textbf{Training}
The model is trained using ADAM \cite{kingma2014adam} optimizer with a learning rate of $5 \times 10^{-6}$ for 60 epochs with batch size of 32 and $L2$ regularization of $0.0001$. The context and pose features are precomputed. In addition, we augment the data at training time by horizontally flipping the images and sub-sampling the over-represented class to equalize the number of crossing and non-crossing samples. 

\section{Experiments}
\textbf{Dataset.}
There are not many datasets suitable for the purpose of pedestrian crossing prediction. One such dataset is JAAD \cite{Rasouli_2017_ICCV_Workshops} which contains videos of pedestrians prior or during crossing. Unfortunately, the number of samples in this dataset is small, no vehicle information is available and sequences are short snippets which are not suitable for long-term predictions. 

For the purpose of this work, we used our newly collected pedestrian intention estimation (PIE) dataset\footnote{\url{http://data.nvision2.eecs.yorku.ca/PIE\_dataset/}}. The dataset comprises 1842 pedestrian tracks captured using an on-board monocular camera while driving in urban environments with various street structures and crowd densities. The samples represent people who are close to the curbs or are at intersections and may or may not have the intention of crossing, e.g. waiting for a bus. Overall, the ratio of non-crossing to crossing events is 2.5 to 1. All video sequences are collected during daylight under clear weather conditions. The videos are continuous allowing us to observe the pedestrians from the moment they appear in the scene until they go out of the field of view of the camera.

The dataset contains bounding box annotations for pedestrians as well as vehicle sensor data such as speed and heading angle per frame. For each pedestrian sample we identified an event point. For those who cross in front of the ego-vehicle, the event is the moment they start crossing. For other samples, the events are set at the time when the pedestrians go out of the field of view of the camera. We randomly split the data into train-test sets with ratio of 60-40 respectively.

\textbf{Metrics.}
As in \cite{Suzuki_2018_CVPR}, we report all the evaluation results using the following metrics: \textit{Accuracy}, \textit{F1} score, \textit{precision} and \textit{recall}. We also use \textit{Area Under Curve (AUC)} metric which, in the case of binary event anticipation, reflects the balanced accuracy of the algorithms.

\subsection{Evaluations}

\textbf{Predicting crossing event.}
We evaluate the performance of our proposed algorithm, stacked with multilevel fusion GRU (\textbf{\textit{SF-GRU}}), against the following models:\\ 
\textbf{\textit{Static.}} This model is inspired by \cite{Rasouli_2017_ICCV_Workshops} and has two VGG16 branches (without \textit{fc} layers and with a global pooling layer) pretrained on ImageNet. One network processes the local context corresponding to the pedestrian crop $c^m_p$  and the other processes the surroundings $c^m_s$ at the last frame of the observation. The outputs of both networks are combined and fed into a \textit{fc} layer for the final prediction.\\
\textbf{\textit{GRU.}} A single-layer GRU \cite{chung2014empirical} trained and tested only on pedestrians' appearances $C_p$ and their surroundings $C_s$. We also use this model with all sources of information which are concatenated and fed into the network at the same time.\\
\textbf{\textit{Multi-stream GRU (M-GRU)}}. Following the approach in \cite{bhattacharyya2018long}, this architecture processes different types of features separately using different GRUs, and feeds the concatenation of the last hidden states of all GRUs into a dense layer for prediction.\\
\textbf{\textit{Hierarchical GRU (H-GRU)}}. This model has a hierarchical structure similar to \cite{du2015hierarchical}. H-GRU processes each feature type using a separate GRU, concatenates the hidden states of all units and then feeds them into another GRU whose last hidden state is used for prediction.\\
\textbf{\textit{Stacked GRU (S-GRU)}}. This is a five-level stacked GRU architecture as described in \cite{yue2015video} which receives the feature inputs at the bottom layer. The inputs to the subsequent GRUs in the higher levels are the hidden states of the GRUs in the previous layers.\\

All evaluations are done on observation sequences of 0.5s (15 frames) duration. The samples are selected with 2s time to event (TTE), the minimum time within which pedestrians make crossing decision according to \cite{Schmidt2009}. 

The results are summarized in Table \ref{table_crossing}. We can see that using the visual information of the local context, even as a single image in the static method can lead to approximately $60\%$ accuracy which can be improved by $9\%$ by performing temporal reasoning using a GRU. 

Using all sources of information, the proposed algorithm \textit{SF-GRU} performs best on all metrics except recall. For this metric single-layer \textit{GRU} performs slightly better (by $1.2\%$) at the expense of more than $6\%$ drop in precision. In addition, the results show that  no performance improvement is achievable by simply adding layers to the network or separating the processing of features with different modalities. 

\begin{figure}
\centering
\includegraphics[width=1\columnwidth]{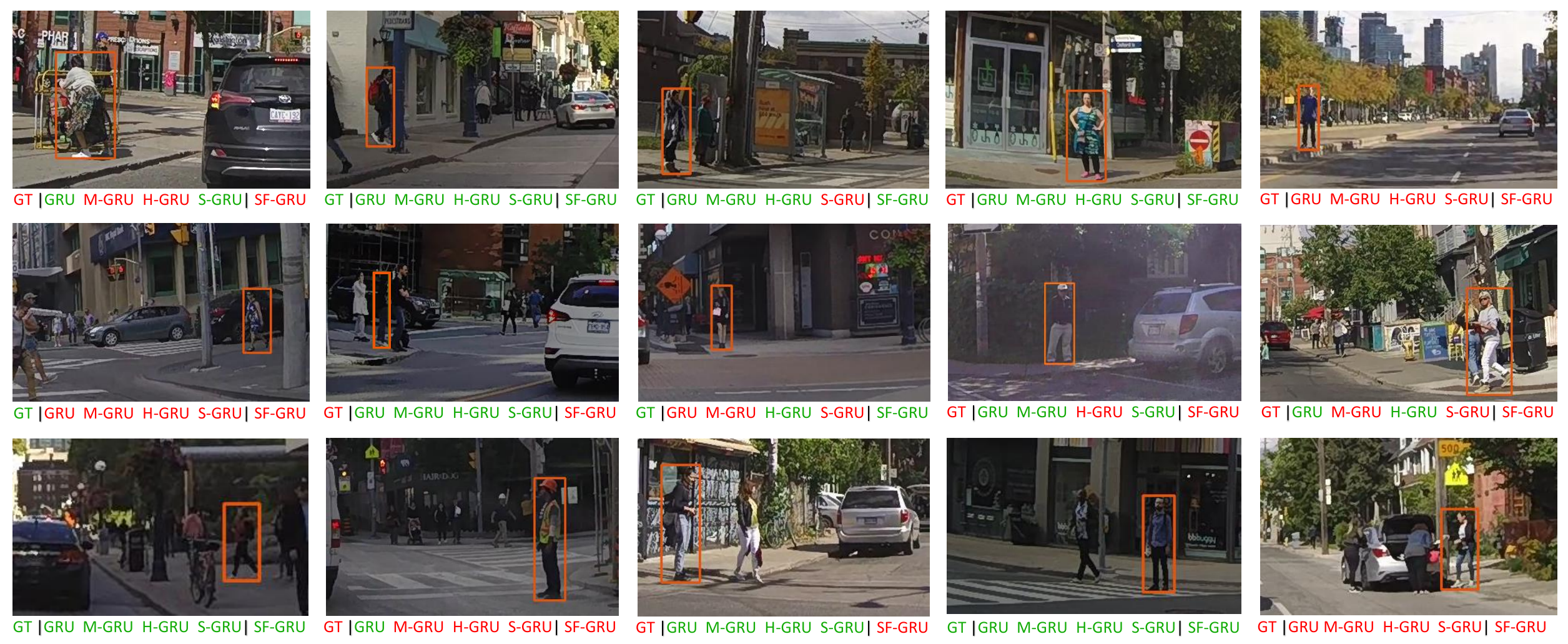}
\caption{Examples of the predictions produced by the proposed algorithm \textit{SF-GRU} and top competing methods, namely \textit{GRU}, \textit{M-GRU}, \textit{H-GRU}, and \textit{S-GRU}. In the examples, \textit{GT} stands for ground truth and  \textit{green} and \textit{red} colors indicate whether the pedestrian will cross in front of the ego-vehicle or not respectively. The instances where the color of the algorithm labels matches the GT means that their predictions are correct.}
\label{results}
\end{figure}

\begin{table}[t]
\centering
\scalebox{0.9}{
\begin{tabular}{|l|c|c|c|c|c|c|}
\hline
 \textit{Models}& \textit{Features} & \textit{Acc} & \textit{AUC} & \textit{F1} & \textit{Prec} & \textit{Recall}  \\ \hline \hline
\multicolumn{1}{|l|}{Static}		&$c_p^m,c_s^m$    &0.592&	0.589&	0.419&	0.328&	0.582\\ 
\multicolumn{1}{|l|}{GRU}			& $C_{p},C_{s}$	  &0.681&	0.644&	0.475&	0.407&	0.570 \\ 
\noalign{\global\arrayrulewidth=0.5mm}
\hline
\noalign{\global\arrayrulewidth=0.1mm} 
\multicolumn{1}{|l|}{GRU}										&$C_p,C_s,P,B,S$     & 0.811& 0.812& 0.685&	0.593&	\textbf{0.812} \\ 
\multicolumn{1}{|l|}{M-GRU}										&$C_p,C_s,P,B,S$	 & 0.804& 0.792& 0.665&	0.585&	0.770 \\ 
\multicolumn{1}{|l|}{H-GRU}										&$C_p,C_s,P,B,S$     & 0.819& 0.805& 0.685&	0.612&	0.776 \\ 
\multicolumn{1}{|l|}{S-GRU}										&$C_p,C_s,P,B,S$ 	 & 0.801& 0.770& 0.643&	0.588&	0.709 \\ 
\multicolumn{1}{|l|}{\textbf{SF-GRU (ours)}}					&$C_p,C_s,P,B,S$     & \textbf{0.844}& \textbf{0.829}& \textbf{0.721}&\textbf{0.657}&	0.800 \\\hline
\end{tabular}
}
\caption{Evaluation results of the algorithms using observation length of 0.5s and time to event (TTE) of 2s. Abbreviations in \textit{features} column are: pedestrian appearance $C_p$, surround context $C_s$, pose $P$, bounding box $B$, and ego-vehicle speed $S$. $c_p^m$ and $c_s^m$ stand for appearance and surround context in the last observation frame respectively.}
\label{table_crossing}
\end{table}

\textbf{When to predict a crossing event.}
The prediction of crossing events may vary depending on TTE as the scene dynamics changes, in particular, when the ego-vehicle motion impacts the way people make a crossing decision. Here we examine the prediction ability of the temporal algorithms with respect to TTE. We alter TTE points from 0s to 3s with steps of approximately 0.16s, a total of 19 different points. To maintain the consistency of data across different time frames, we only sample from pedestrian tracks equal to or longer than 3.5s (the maximum TTE time in the experiment + observation length). All other parameters including the observation sequence length remain the same as before.

\begin{figure}
\centering
\subfigure[]{
\includegraphics[width=0.40\textwidth]{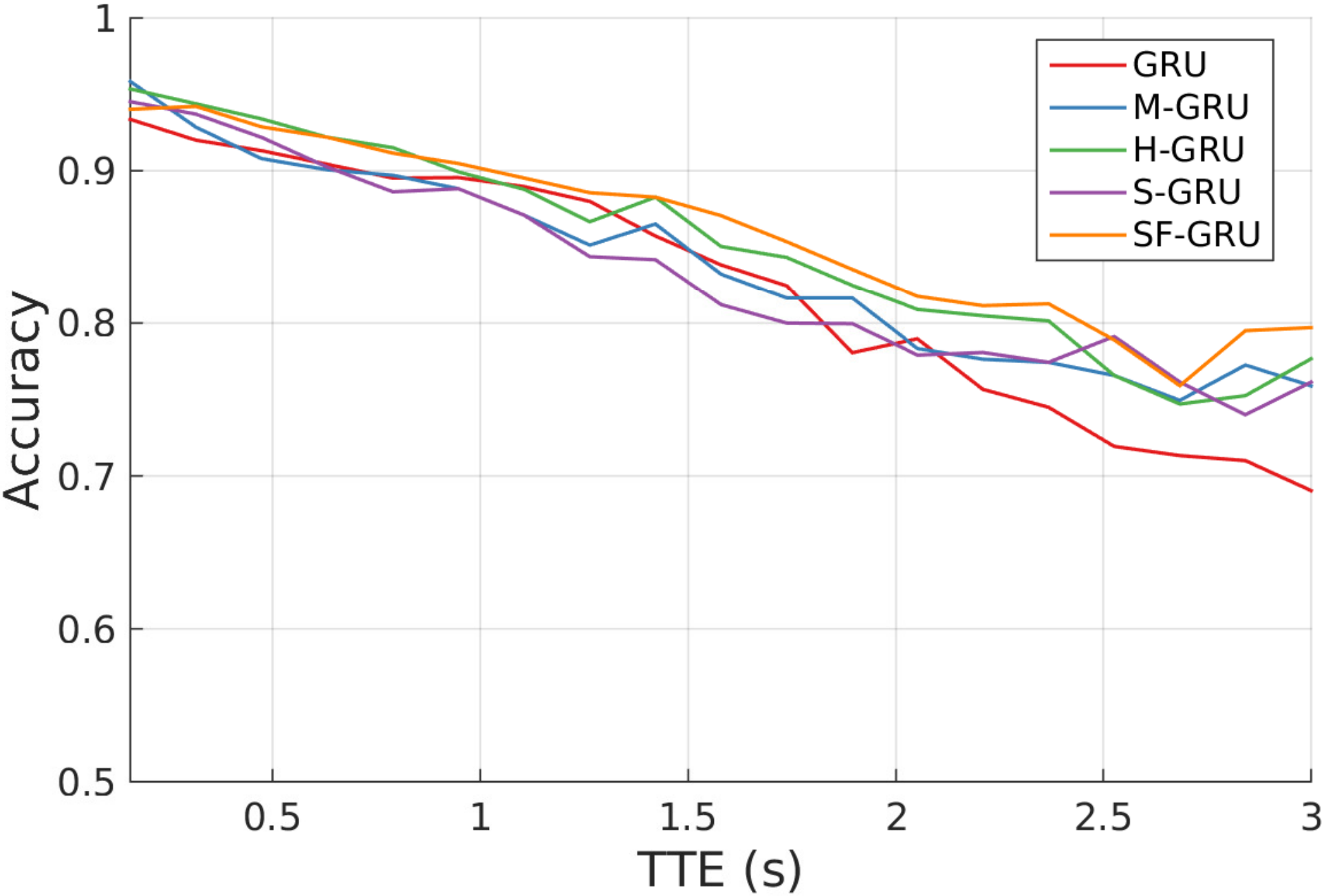}
\label{model_vs_tte}
}
\hspace{1.5cm}
\subfigure[]{
\includegraphics[width=0.40\textwidth]{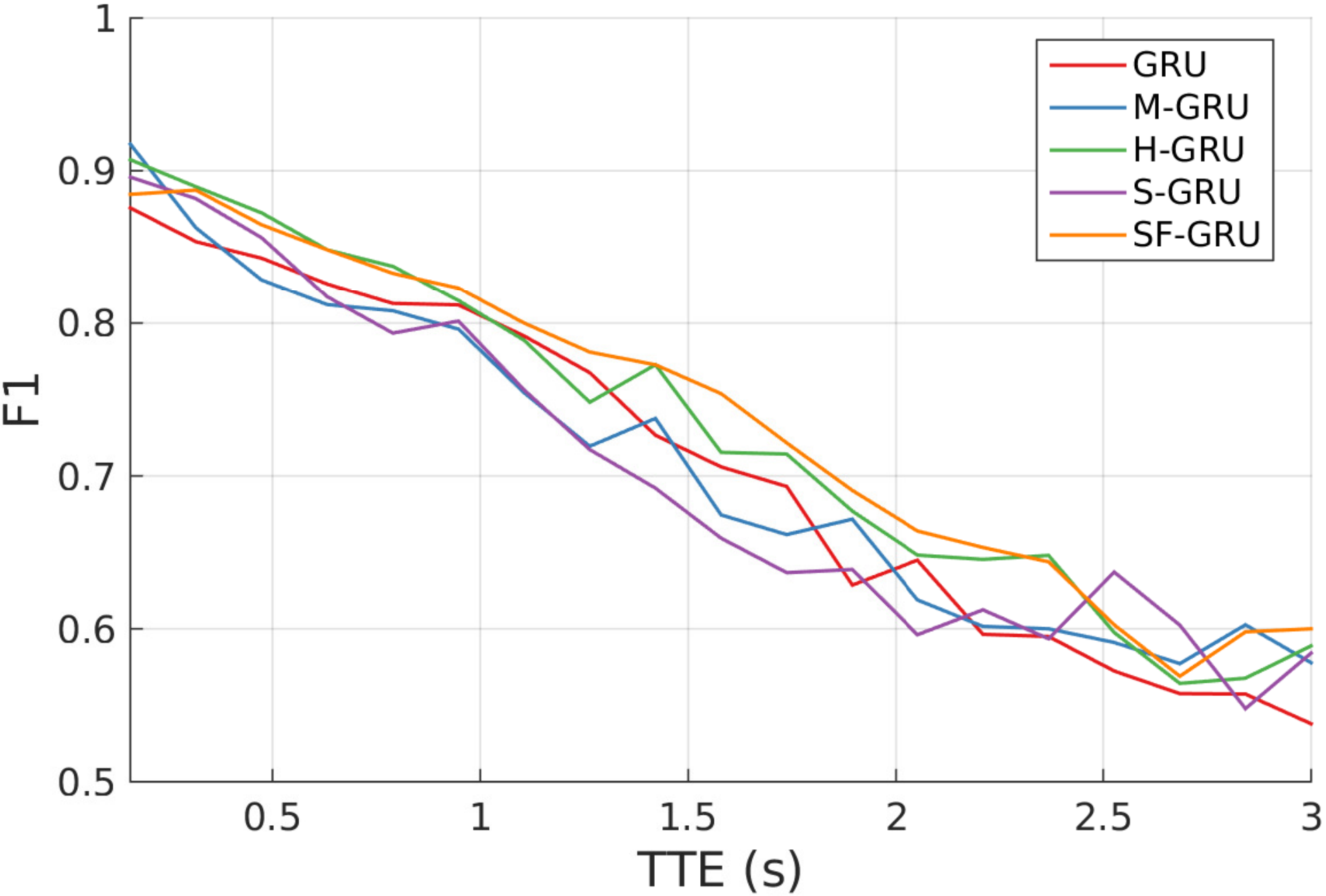}
\label{tte_vs_obs}
}
\caption{The performance of the algorithms with respect to varying time to event (TTE) points with 0.5s observation length.}
\label{tte}
\end{figure}

The proposed algorithm \textit{SF-GRU} performs best for the most part at different TTE points (see Figure \ref{tte}). At early TTE times where the intention of pedestrians becomes obvious, all algorithms perform similarly well. However, as expected, the performance of the algorithms degrades gradually (some at a faster rate than others) as the observations are moved  further away from the time of the event. We can also see that the single-layer \textit{GRU} only performs better than \textit{M-GRU} and \textit{S-GRU} up to 2s TTE after which its performance drops rapidly.

\begin{figure}[t]
\centering
\includegraphics[width=1\textwidth]{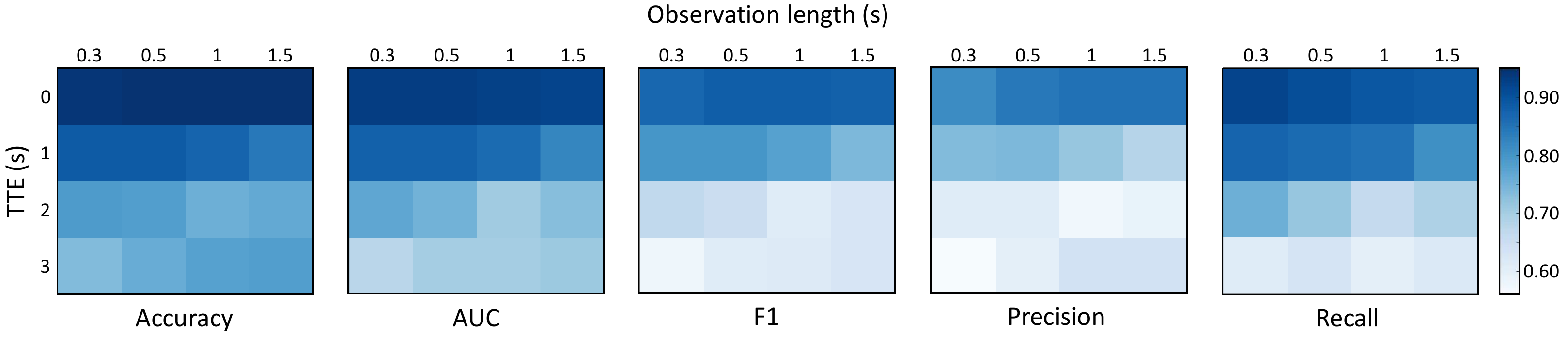}
\caption{The changes in the performance of \textit{SF-GRU} according to varying observation length and time to event (TTE).}
\label{tte_obs}
\end{figure} 

\textbf{The effect of observation length on prediction.}
Longer observation time can potentially provide more information but at the same time may add noise. We examine the effect of observation length on the proposed algorithm \textit{SF-GRU} with respect to different TTE points. For the same reason as mentioned in the previous experiment, we only sample from tracks with length equal to or longer than 4.5s (the longest observation length + the largest TTE value). In total, we examine 16 different combinations.

As shown in Figure \ref{tte_obs}, on most metrics the improvement gain is only on samples very close (0s) or far away (3s) from the event. In these cases, precision can be improved by longer observations at the expense of reducing the recall. In critical decision regions of 1-2s, however, a small gain is achieved by increasing observation from 0.3s to 0.5s after which point the performance drops rapidly. This could be due to noise in longer observations caused by accumulation of the changes in the scene dynamics. For instance, within 1.5s observation window, the speed of the vehicle can change significantly which can have a considerable effect on predicting pedestrian crossing behavior.

\textbf{Feature types and prediction accuracy.} We examine the contribution of each feature type on the performance of the proposed algorithm. In addition to the features discussed earlier, we also evaluate two other types of features: displacement $D$ (the center coordinates of the bounding boxes) and full context $C_{p+s}$ which is the pedestrian appearance and surround context in a single frame, not as decoupled features as proposed earlier.

As shown in Table \ref{features_accuracy}, we can see that adding contextual information in addition to pedestrian appearance to the network improves the overall performance by more than $18\%$. We also see that decoupling appearance and surround context boosts the accuracy by almost $3\%$ owing to precision gain. Another observation is that using bounding box coordinates instead of center coordinates improves the results by $2\%$. This can be due to the fact that the changes in the scale of the bounding boxes in a sequence can add another layer of information, e.g. the movement of pedestrian or the changes in their distance to the ego-vehicle.

\begin{table}[]
\centering
\scalebox{0.9}{
\begin{tabular}{|l|c|c|c|c|c|}
\hline
\textit{Features} & \textit{Acc} & \textit{AUC} & \textit{F1} & \textit{Prec}& \textit{Recall}\\   
\hline \hline  						
 $C_p$		    &0.660&	0.622&	0.448&	0.380&	0.546\\		
 $C_{p+s}$		&0.666&	0.650&	0.483&	0.397&	0.618\\		
 $C_p,C_s$		&0.692&	0.645&	0.475&	0.417&	0.552\\		
 $C_p,C_s,P$	&0.745&	0.705&	0.554&	0.498&	0.624\\		
 $C_p,C_s,P,D$	&0.796&	0.765&	0.636&	0.580&	0.703\\		
 $C_p,C_s,P,B$	&0.816&	0.781&	0.661&	0.619&	0.709\\
 $C_p,C_s,P,B,S$&\textbf{0.844}&	\textbf{0.829}&	\textbf{0.721}&	\textbf{0.657}&	\textbf{0.800}\\
\hline
\end{tabular}
}
\caption{The impact of different sources of information on the performance of \textit{SF-GRU}. The feature types are as follows: $C_p$ pedestrian context (appearance), $C_s$ surround context, $C_{p+s}$ full context , $P$ pose, $D$ displacement (center coordinates), $B$ bounding box, and $S$ speed.}
\label{features_accuracy}
\end{table}

\textbf{The order of fusion and performance.}
In this experiment, we investigate how different fusion strategies alter the performance. Since reporting on all possible permutations of different sources of information is prohibitive, we only include a subset of these permutations to show the fluctuations in the overall performance.

A summary of the results is provided in Table  \ref{fusion_accuracy}. Here, it is shown that when more complex features such as local context are infused into higher levels of the network, the performance gets worse. By inputting different feature types in the right order, that is by moving  simpler features, such as speed, to the higher levels of the stack, the performance improves by up to $9\%$ on accuracy, $10\%$ on recall and more than $15\%$ on precision.  This can be due to the fact that more complex visual features, which benefit more from deeper spatial analysis, are inputted at the bottom layers of the network while simpler features such as trajectory coordinates are entered at the higher levels.
\begin{table}[]
\centering
\scalebox{0.9}{
\begin{tabular}{|l|c|c|c|c|c|}
\hline
\textit{Features} & \textit{Acc} & \textit{AUC} & \textit{F1} & \textit{Prec}& \textit{Recall}\\   
\hline \hline  						
 $P,S,B,C_p,C_s$		    &0.753&	0.737&	0.590&	0.509&	0.703\\		
 $S,B,C_p,C_s,P$			&0.784&	0.759&	0.624&	0.557&	0.709\\		
 $B,C_p,C_s,P,S$			&0.798&	0.776&	0.647&	0.579&	0.733\\		
 $S,C_p,C_s,P,B$			&0.810&	0.785&	0.661&	0.602&	0.733\\		
 $C_p,B,C_s,S,P$			&0.813&	0.803&	0.679&	0.619&	0.788\\		
 $C_p,C_s,P,B,S$			&\textbf{0.844}&	\textbf{0.829}&	\textbf{0.721}&	\textbf{0.657}&	\textbf{0.800}\\
\cline{1-6}
\end{tabular}
}
\caption{Feature fusion strategies and their impact on the performance of the proposed algorithm \textit{SF-GRU}. The feature types are as 
follows: $C_p$ pedestrian context (appearance), $C_s$ surround context, $P$ pose, $B$ bounding box, and $S$ speed.}
\label{fusion_accuracy}
\end{table}
\section{Conclusion}
We presented a novel stacked RNN architecture in which different sources of information including pedestrian and the vehicle dynamics, pedestrian appearance and their surroundings are fused gradually at different levels of processing. Using empirical evaluations we showed that our approach performs best compared to alternative RNN architectures. In addition, we demonstrated how different sources of information and data fusion strategies into the network can impact crossing action prediction. We highlighted that the performance is optimal when more complex features are fed to the bottom layers of the network and the simpler ones at the higher levels. Although the proposed architecture was presented in the context of pedestrian crossing prediction,  other applications of similar nature e.g. activity recognition may also benefit from using this approach.

\section*{Acknowledgements}
This work was supported by the Natural Sciences and Engineering
Research Council of Canada (NSERC), the NSERC Canadian
Robotics Network (NCRN), the Air Force Office for Scientific Research (USA), and the Canada Research Chairs Program through grants to JKT.

\bibliography{bmvc}
\end{document}